\documentclass[conference]{IEEEtran}
\IEEEoverridecommandlockouts
\usepackage{cite}
\usepackage{amsmath,amssymb,amsfonts}
\usepackage{algorithmic}
\usepackage{graphicx}
\usepackage{textcomp}
\usepackage{xcolor}
\usepackage{tabularx}
\def\BibTeX{{\rm B\kern-.05em{\sc i\kern-.025em b}\kern-.08em
    T\kern-.1667em\lower.7ex\hbox{E}\kern-.125emX}}
\begin{document}

\title{PatrolVision: Automated License Plate Recognition in the wild\\}

\author{\IEEEauthorblockN{1\textsuperscript{st} Anmol Singhal}
\IEEEauthorblockA{\textit{New York University} \\
as15151@nyu.edu}
\and
\IEEEauthorblockN{2\textsuperscript{nd} Navya Singhal}
\IEEEauthorblockA{\textit{University of Texas} \\
navya.singhal@utexas.edu}
\and
}

\maketitle

\begin{abstract}
Adoption of AI driven techniques in public services remains low due to challenges related to accuracy and speed of information at population scale. Computer vision techniques for traffic monitoring have not gained much popularity despite their relative strength in areas such as autonomous driving. Despite large number of academic methods for Automatic License Plate Recognition (ALPR) systems, very few provide an end to end solution for patrolling in the city. This paper presents a novel prototype for a low power GPU based patrolling system to be deployed in an urban environment on surveillance vehicles for automated vehicle detection, recognition and tracking. In this work, we propose a complete ALPR system for Singapore license plates having both single and double line creating our own YOLO based network. We focus on unconstrained capture scenarios as would be the case in real world application, where the license plate (LP) might be considerably distorted due to oblique views. In this work, we first detect the license plate from the full image using RFB-Net and rectify multiple distorted license plates in a single image. After that, the detected license plate image is fed to our network for character recognition. We evaluate the performance of our proposed system on a newly built dataset covering more than 16,000 images. The system was able to correctly detect license plates with 86\% precision and recognize characters of a license plate in 67\% of the test set, and 89\% accuracy with one incorrect character (partial match). We also test latency of our system and achieve 64FPS on Tesla P4 GPU 
\end{abstract}


\section{Introduction}
Automatic License Plate Recognition (ALPR) in unconstrained environments plays a central role in many applications as a first crucial step towards many practical and relevant applications such as automatic traffic law enforcement, detection of stolen vehicles or toll violation, traffic flow control, etc. However due to the large variability in image acquisition conditions (illumination, capture angle, distance from camera, etc.), blurry images, and variability of license plate numbers, which vary from one country to another, this task is still an open problem. In this paper, we collect a new dataset in Singapore (sample images in Fig 1) and propose a benchmark for ALPR systems. The system is designed to be deployed for real time vehicle recognition and tracking using dashcams on police vehicles and the onboard edge systems. This system will be used to monitor the flow of traffic, vehicle identification, and violation tracking. Automated License plate detection and recognition (ALPR) is one of the key aspects of an automated patrolling system. The advantage ALPR has is high accuracy, eliminates human intervention, and does not require any additional hardware than already present on our city streets/vehicles, as compared to the Ultra High Frequency—Radio Frequency Identification (UHF-RFID) systems \cite{RFID}. 

\begin{figure*}[htbp]
\centering
\includegraphics[width=\textwidth,keepaspectratio]{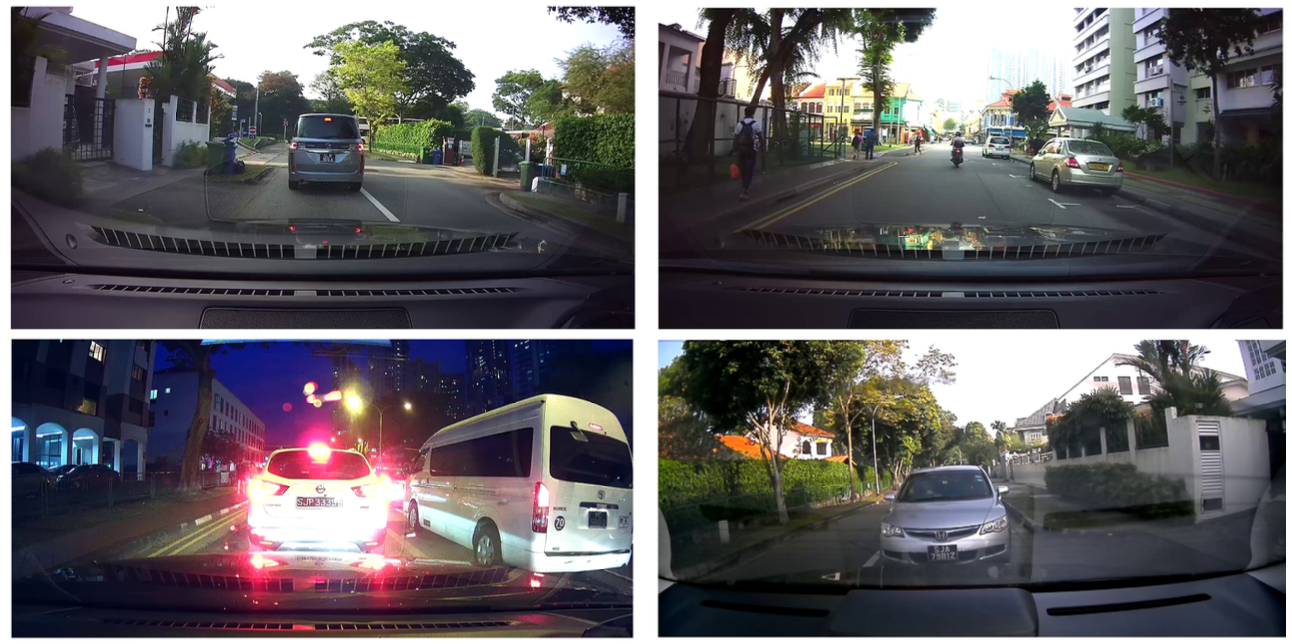}
\caption{Examples of real world images captured from dash cams}
\label{fig}
\end{figure*}

\begin{figure*}[htbp]
\centering
\includegraphics[width=\textwidth,keepaspectratio]{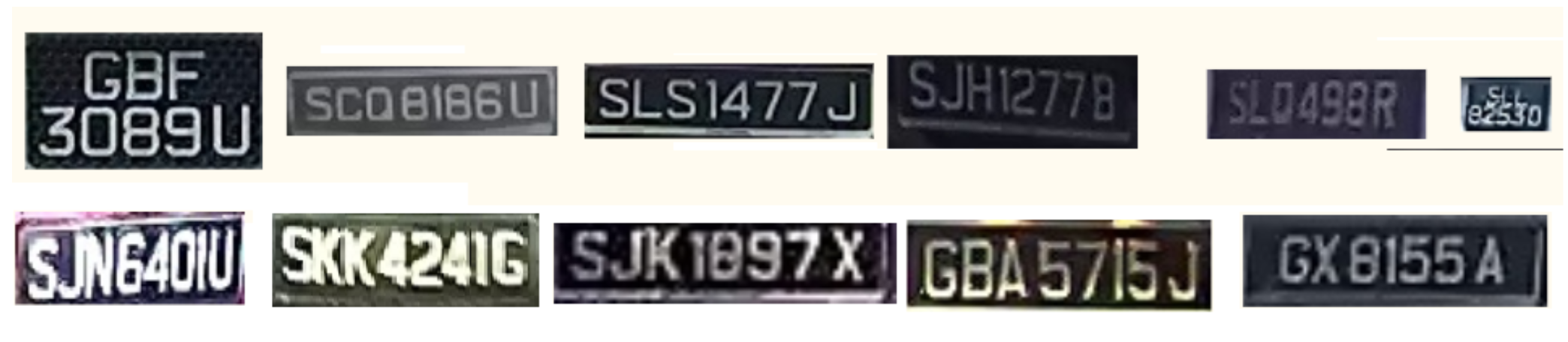}
\caption{Examples of detected license plates from the full captures}
\label{fig}
\end{figure*}

Vision-based ALPR itself can be broken into multiple sub-tasks \cite{BrazilLP}: License Plate Detection (LPD), License Plate Segmentation (LPS),  Character Recognition (CR), and sequence reconstruction. The objective is to accurately locate and recognize license plates from images or video streams, which presents challenges due to varying lighting conditions, plate sizes, orientations, and environmental factors like occlusions and distortions. License plate detection is defined as being able to draw a bounding box around the license plate on multiple vehicles from images captured in the wild. License plate segmentation and character recognition relate to being able to accurately segment the characters and then identify them. Sequence reconstruction becomes important for this task since the information is highly dependent on the ordering of the characters. Some existing methods are focused on combining these three subtasks \cite{2,3,4} and some on only one or two subtasks \cite{1, 5, 7}. In computer vision research, the combined subtasks; LPS and CR are related to Optical Character Recognition (OCR) \cite{8,9,10,11}.

In this paper, we propose the ALPR system by combining LPD and CR without detecting the related vehicle. The proposed method is based on Deep Convolutional Neural Network. Recent studies proved the effectiveness and superiority of Convolutional Neural Networks in many Computer Vision tasks such as image classification, object detection and semantic segmentation. However, running most of them on embedded devices remains a challenging problem. In this paper, we demonstrate our lightweight model achieves state-of-the-art accuracy while being fast enough to be deployed in a real-time scenario on low-power GPUs. To achieve the throughput, we improve standard non-max suppression to reduce kernel calls and achieve 38 FPS for LPD and 68 FPS for CR on NVidia 1080Ti. Also, our model is real-time (at 7.7 FPS) on Jetson TX2 with high accuracy on challenging license plates. Our main contributions can be summarized as follows:

\begin{itemize}
\item Our dataset is collected from videos collected from a camera mounted on the dashboard of a car in Singapore. The processing is frame-based; each video frame is processed individually and results are returned for individual frames (no video tracking). Application of the proposed method to real traffic surveillance video shows that our approach is robust enough to handle difficult cases, such as perspective and camera-dependent distortions, hard lighting conditions, change of viewpoint, etc. The dataset will be made available for academic research on request.
\item Benchmark various object detectors for LPD
\item Design a novel YOLO-based network for LPR
\item Profile and build the system to be used in real-time patrolling scenario

\end{itemize}

The rest of this paper is arranged as follows: Section 2 discusses the applications and techniques of existing ALPR systems. Section 3 presents the proposed system in detail. Section 4 describes and discusses the experimental results, followed by the conclusions in the last section.

\section{Related works}

This section discusses some deep learning-based object detection methods which can be used to detect License plates. Some computer vision and pattern recognition methods are also revised which are commonly used for ALPR.

\subsection{Object detection}

Since the introduction of neural networks, detection frameworks have become increasingly fast and accurate such as Faster RCNN \cite{17}, YOLO \cite{14}, SSD \cite{16}, RetinaNet \cite{24}, RefineDet \cite{25}, RFBNet \cite{26}. Faster R-CNN uses a region proposal network to create boundary boxes and utilizes those boxes to classify objects. While it is considered the start-of-the-art in accuracy, this approach has been too computationally intensive for embedded systems and, even with high-end hardware, too slow for real-time applications. SSD speeds up the process by eliminating the need for the region proposal network. As mentioned in \cite{15}, SSD with a 300×300 input size significantly outperforms its 448×448 YOLO counterpart in accuracy and speed. SSD is based on a feed-forward convolutional network that produces a fixed-size collection of bounding boxes and scores for the presence of object class instances in those boxes, followed by a non-maximum suppression step to produce the final detections. The concept of RFBNet \cite{26} is inspired by receptive fields in humans and strengthens deep features from lightweight CNNs to improve speed and accuracy in a detector. RFB makes use of multi-branch pooling with varying kernels corresponding to RFs of different sizes, applies dilated convolution layers to control their eccentricities, and reshapes them to generate the final representation. We then assemble the RFB module to the top of SSD, a real-time approach with a lightweight backbone, and construct an advanced one-stage detector (RFB Net). Additionally, the RFB module is generic and imposes few constraints on the network architecture.

\subsection{Automatic license plate recognition}
Existing methods for ALPR are focused on the subtasks; LPD, to detect and localize the license plate; LPS, to segment out the individual characters from the background after the LPD; and finally, CR, to recognize all the segmented characters and return the full plate string.

\subsubsection{License Plate Detection (LPD)}
License plate detection is the foundation of license plate recognition, and its accuracy directly affects the results of character recognition in license plates. Currently, there are two main categories of license plate detection methods in common use. One is direct license plate detection in which given the input image we can predict the location, height and width information of the license plate. In indirect license plate detection, under the complex environment, researchers take advantage of the prior knowledge between the license plate and the car body, such as the position relationship between the rear lights and the license plate. Since the license plate is part of the body of a car, human experience can determine the approximate location of the license plate even if it cannot be located immediately at first glance. 

Further, the techniques to solve both problems can be classified into traditional and deep learning-based techniques. In traditional techniques, researchers rely on the intrinsic properties of license plates such as edges, colors, local textures, and morphological analysis as manual image features for license plate detection. Given sufficient training data, deep learning-based license plate detection algorithms have powerful feature representation and high performance compared to license plate detection algorithms based on traditional methods. They can usually be divided into two categories: one-stage detection methods and two-stage detection methods. Most of the existing ALPR system detect the car and then detect the L, aiming to reduce the search region and the number of false positives. In \cite{5,6,18} proposed LPD using Histogram of Oriented Gradients (HOG) as feature descriptors and Support Vector Machine (SVM) for classification. For European and USA license plates, OpenALPR \cite{19} used Local Binary Patterns as a feature descriptor to detect LP. Three deep CNNs are used to perform LPD, LPS and CR, respectively in \cite{2} which outperform OpenALPR. However, the network architecture and training procedures are not given as this method is a commercial method.

\subsubsection{License Plate Recognition}

Traditional license plate character recognition methods have two main steps: license plate character segmentation and character recognition. Segmentation of license plate characters is to separate all characters of a license plate before character recognition in order to match the input of the character recognition algorithm. The usual character segmentation methods are character detection, concatenated domain search, and vertical projection. The segmented individual characters are then fed into the recognition module to obtain the recognition results. Character recognition methods usually include template matching, feature statistics, and machine learning.

To segment the characters from LP, firstly the foreground is extracted from the background by using some existing binarization methods. After that the connected components are extracted to identify the characters with bounding boxes \cite{20}. But if the characters are connected then this method cannot segment out the characters. So to separate the connected characters a processing method is required to perform after the second step. Moreover, due to textured background, low contrast between the background and the characters, or non-even illumination the binarization methods do not perform well. Some methods, extract the Extremal Regions \cite{8} and then segment out the text regions \cite{9,10}.

Some existing methods \cite{18} proposed a learning approach based on SVM to recognize characters. Restricted Boltzmann Machines (RBM) is also used for CR and it is better than SVM \cite{12,21}. In \cite{21}, the author used HoG features with RBM to recognize Chinese characters. Hinton et.al. \cite{12} proposed Deep Belief Networks which is similar to RBM. Some deep learning methods \cite{2,3} are also used to recognize characters. In [3], CR is performed by 16 layer CNN based on Spatial Transformer Networks.

\subsubsection{End to end ALPR}

Some recent works have proposed a License Plate Pipeline where they combined all the subtasks of ALPR with deep learning techniques. In the paper \cite{30}, the author performed both detection and recognition with the YOLO framework. This method first detects the car and then detects the LP on the car region, which compromises its execution time. In the paper \cite{13}, the author improved the accuracy by separating the recognition tasks into segmentation and classification. Hsu et al. \cite{31} target LP detection only based on the YOLOv2 architecture. They do not handle the license plate recognition. Li et al. \cite{32} also applied a region proposal network (RPN) unifying detection and recognition in a single network. However, this method is not real-time (3.4 FPS as reported). Most of these works are not for both single and double line LP \cite{BrazilLP, chinaalpr} and some do not perform in real-time.

\section{Proposed method}

In this section, we describe our network architecture design in detail. The block diagram of our ALPR system is shown in Fig. 3. We target the ALPR system into four main subtasks: the first to perform the LPD and the second to perform the combination of LPS and CR, we term this combination License Plate Recognition(LPR). The third subtask is to use heuristic rules to improve recognition accuracy and the final step is to arrange the characters in a proper sequence.

\begin{figure*}[htbp]
\centering
\includegraphics[width=\textwidth,keepaspectratio]{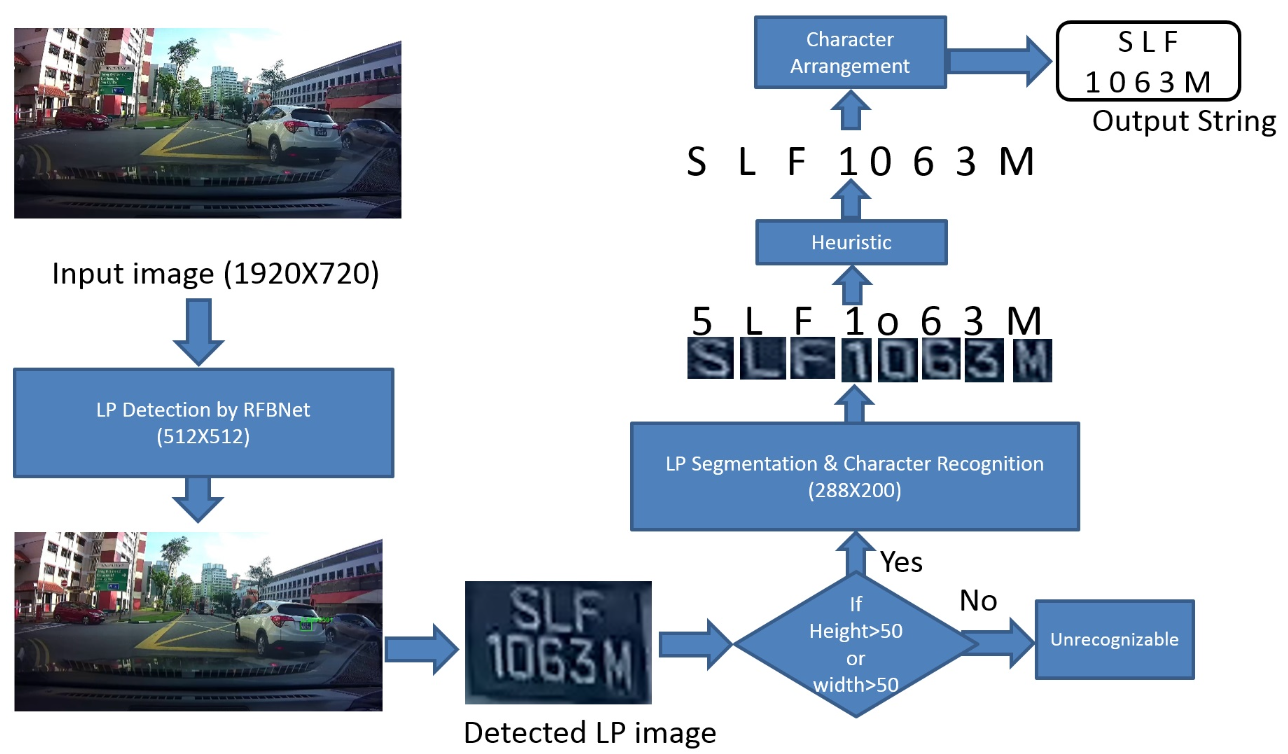}
\caption{Block-diagram of the proposed methodology}
\label{fig}
\end{figure*}

\subsection{License plate detection}

For the first stage, we consider the LPD as an object detection problem with an image size of 1920 × 720. We target some of the state-of-the-art object detectors based on the Deep learning approach. We evaluate the accuracy and speed on 6 state-of-the-art object detectors(Faster RCNN, SSD, YOLO, RetinaNet, RefineDet and RFBNet) for LPD on our in-car dataset as shown in Table1. Based on this experiment, we find that RFBNet delivers the highest accuracy with the fastest speed just before RefineDet.

\begin{table*}[ht]
\centering
\begin{tabularx}{\textwidth}{|c|X|X|X|X|X|X|}
\hline
 & Faster R-CNN & SSD & YOLO & RetinaNet & RefineDet & RFBNet \\ 
\hline
Precision & Row 1 & 73 & 79 & 80 & 84 & 86 \\ 
\hline
Speed (FPS) & 12 & 27 & 32 & 19 & 29 & 35 \\
\hline
Backbone & Resnet 101 & VGG16 & DarkNet53 & ResNet50 & VGG16 & VGG16 \\
\hline
Input size & 600x600 & 512x512 & 608x608 & 600x600 & 512x512 & 512x512 \\
\hline
\end{tabularx}
\caption{Comparison of Object Detection Models for license plate detection on our dataset}
\end{table*}

\subsection{License plate recognition}

After detecting the LP, the characters should be segmented and recognized. But the unconstrained nation of our dataset presents a unique challenge. The majority of the license plates detected by the first stage end up being smaller than 50 pixels height or width which is challenging even for our annotators to read and annotate. To identify the noisy LP before passing to the second stage, we filter out LP with a height or width of less than 50 pixels. These detected license plates tend to be ones captured in the background when there are multiple cars in the frame. Since we tackle ALPR frame by frame, we can recognize these vehicles when our camera moves closer. After filtering the LP based on size we pass the LP images to the second stage.

For the second step, we use a deep learning-based object detector; YOLO \cite{14} to detect and recognize the characters. We use 0-9 digits and A-Z alphabets (only capital letters are allowed in LP) to classify the characters. As the digit 0 and alphabet O have similar shapes, we take these as one class. So overall we have 35 classes. The YOLO has a fixed input and output aspect ratio as 1:1 which detects objects in both portrait and landscape images. However as the LP image is in rectangular in shape, 1:1 aspect ratio as input is not suitable. In the paper \cite{13}, the author also changed the aspect ratio of the input image for the YOLO network to 3:1 to detect and recognize characters in single-line LP. The author changed the input size for the YOLO network to 240×80. We mainly follow the architecture used which is also based on YOLO. But we target to detect characters for both single and double line LP which has aspect ratiosof  nearly 3:1 and 3:2 respectively. We tried different input sizes as 160×120, 200×160, and 288×200 to the architecture and 240×80 used in [13] with the pre-trained models on our dataset. It achieves 64.4\%, 63.6\%, 67\% and 56.5\% accuracy respectively. With an input size of 240×80 (used in \cite{13}), it achieves 83\% accuracy for single line LP, but 30\% accuracy for double line LP of our dataset. So overall, it achieves 56.6\% accuracy which is not suitable for multi-line LP images. Therefore, we change the input size(288×200 against 240×80) of the detection network and the aspect ratio close to 3:2. As we use a smaller input than the size for YOLO, we reduce some max pooling layers. We use three max-pooling instead of five. Moreover, the output size of YOLO is 13×13, which is not enough for our application. So we put 36×25 as output size of the network which reduces the chance of losing important information for both one and double-line LP. The final architecture of the proposed network is provided in Fig 4.

\begin{figure*}[htbp]
\centering
\includegraphics[width=\textwidth,keepaspectratio]{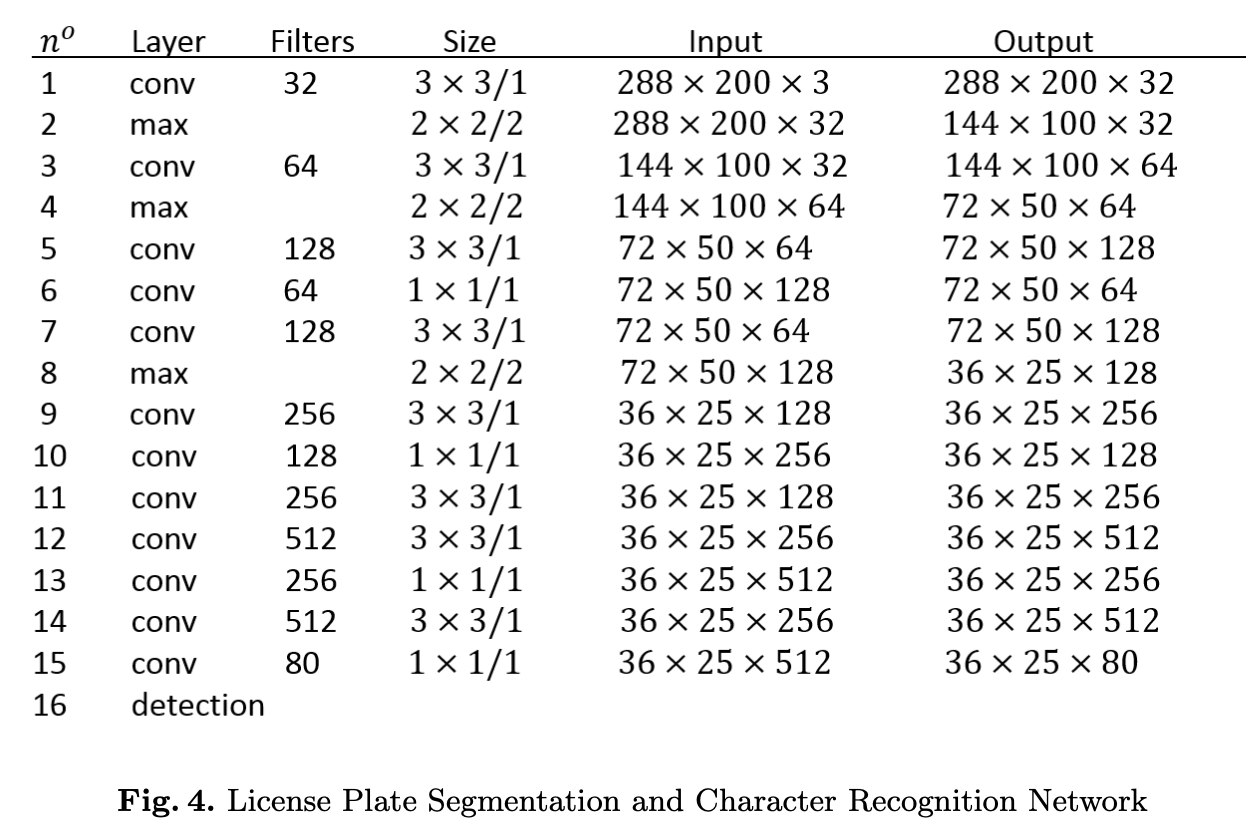}
\caption{License Plate Segmentation and Character Recognition Network}
\label{fig}
\end{figure*}

\subsection{Post processing}

Singaporean license plates have a very specific format for both single and double line \cite{29} which can be used as heuristic rules to improve recognition accuracy. The layout is shown in Fig. 5 for both single and double-line LP. The layout is mainly divided into four parts;

\begin{figure*}[htbp]
\centering
\includegraphics[width=\textwidth,keepaspectratio]{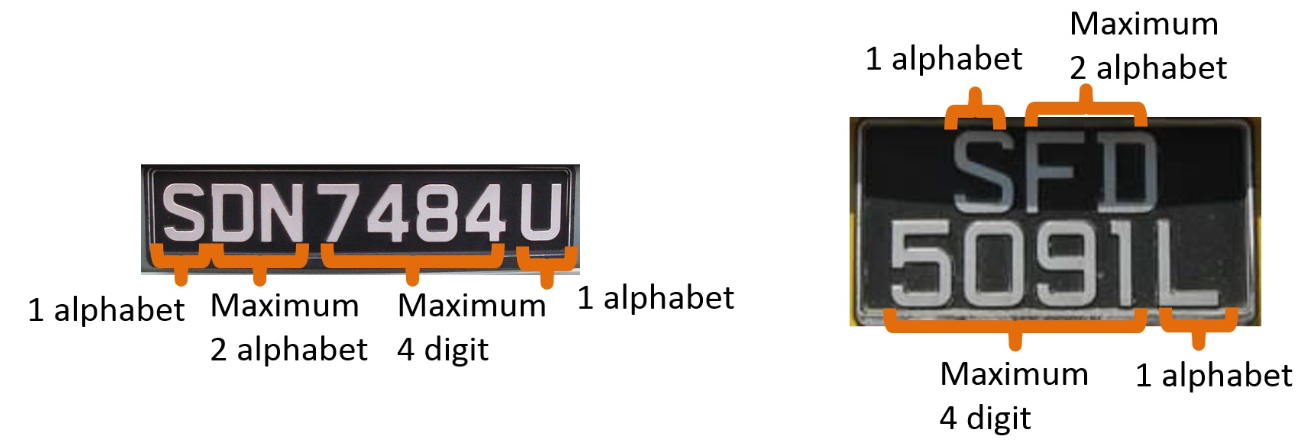}
\caption{Singaporean LP layout for single line and double line: starting with an alphabet followed by maximum 2 alphabet followed by maximum 4 digits and ending with an alphabet}
\label{fig}
\end{figure*}

\begin{itemize}
    \item The first part consists of one alphabet stands for Vehicle class (e.g.“S” stands for a private vehicle since 1984)
    \item The second part consists of upto two alphabets for Alphabetical series (“I” and “O” are not used to avoid confusion with “1” and “0”)
    \item The third part consists of upto four-digit for the Numerical series
    \item The fourth or last part consists of only one alphabet for Checksum letter except “F”, “I”, “N”, “O”, “Q”, “V” and “W”
\end{itemize}

Based on these rules for the layout, we design heuristics to change digit to alphabet and some alphabet to digit which have high similarity index. If the digit is detected in the first, second, or last part of the layout, we convert it to an alphabet that has a similar shape. For example, we change from “5”/“3” to “S”, “8” to “B”, “7”/“2” to “Z”, “4” to “A”, “1” to “T” etc. Similarly, if the alphabet is detected in the third part of the layout, we convert it to a digit that has a similar shape. For example, we change from “S” to “5”, “B” to “8”, “Z” to “7”, “A” to “4”,“ I”/“T”/“L” to “1”, “D”/“O”/“Q” to “0” etc.

\subsection{Character arrangement}

The final step of the proposed method is to return the sequence of characters. As the LP can be a single or double line, we cannot directly sort the characters based on the coordinates detected by the CR network. The network from the second stage gives the classification score and the bounding box with the coordinates of the left-top corner and right-bottom corner of each character. Let's assume that the LP image has height and width as h and w. Let it have 8 characters and the detected bounding box with coordinates of the left-top corner are {(x1,y1), (x2,y2),...,(x8,y8)}. ym and yM are the minimum and maximum values from {y1,y2,...,y8}. Then we decide the category of LP as a single or double line based on equation 1.

\begin{equation}
\textit{Category} =
\begin{cases} 
    \textit{SingleLineLP} & \text{if } (y_M - y_m) < w \times 0.3, \\
    \textit{DoubleLineLP} & \text{otherwise}.
\end{cases}
\end{equation}

For single-line LP, we arrange the characters in increasing order of x-coordinates and return the sequence of characters. For double line LP, we divide the characters into first line and second line based on the equation 2. Then we arrange the characters in increasing order of x-coordinates for both lines individually. Finally, we combine the characters from the first and the second line and return the string.

\begin{equation}
\textit{Line}_i = 
\begin{cases} 
    \textit{FirstLine}, & \text{if } y_i < w \times 0.3, \\
    \textit{SecondLine}, & \text{otherwise},
\end{cases}
\quad i = 1, 2, \dots, 8
\end{equation}

\section{Experiments}
In this section, we discuss the dataset construction and evaluate the performance of the proposed system using the dataset.

\subsection{Dataset construction}\label{AA}

We present a new dataset that contains a diverse set of video sequences recorded in street scenes from Singapore, with high quality(1920×1080). We collect videos by two cameras mounting on the front and back dashboard of a car. The videos are collected for 3 months(Oct, Nov, and Dec 2017) with different lighting conditions (day, night, rain, etc.). The videos are split frame by frame with 30 FPS and we sampled for 5 seconds to construct our dataset. We extract overall 16910 frames. We use the images collected in Oct and Nov for training and in Dec for testing LP Detector. The details of the dataset are shown in Table 2.
The dataset is intended for assessing the performance of computer vision tasks for real-world applications in real time. The annotation is done for all the LP from all the collected frames with bounding boxes and sequence of characters. The bounding boxes are used to train and evaluate the CNN model to detect the LP. The sequence of characters of LP is used to evaluate the CR method. Due to the small resolution of some LP images, the characters of those LPs are not recognizable. To reduce false positives, we divide the LP images of our dataset into recognizable and unrecognizable based on LP image resolution. The details of the LP annotation are shown in Table 3.


\begin{table}[ht]
\centering
\caption{Details of In-car Dataset}
\begin{tabular}{|c|c|c|c|}
\hline
\textbf{Category} & \textbf{Month} & \textbf{Number of Images} \\ 
\hline
Training 2017 & Oct, Nov & 13,768 \\ 
\hline
Testing 2017 & Dec & 3,142 \\ 
\hline
\textbf{Total} & & \textbf{16,910} \\
\hline
\end{tabular}
\end{table}

\begin{table}[ht]
\centering
\caption{Details of LP Dataset}
\begin{tabular}{|c|c|c|c|}
\hline
\textbf{Number of Lines} & \textbf{Recognizable} & \textbf{Unrecognizable} & \textbf{Total} \\ 
\hline
Single & 1439 & 699 & 2138 \\ 
\hline
Double & 2149 & 1369 & 3518 \\ 
\hline
\textbf{Total} & \textbf{3588} & \textbf{2068} & \textbf{5656} \\ 
\hline
\end{tabular}
\end{table}

\subsection{Results}

In this section, we evaluate the performance of the proposed ALPR system. First, we perform the LP detector based on RFBNet and we evaluate the detection precision. Second, we perform the CR method and evaluate the recognition accuracy. We use NVidia 1080Ti with 8GB RAM for training the LP detector. For testing LP detector and CR method, we use NVidia Jetson TX2 and inference server NVidia Tesla P4.

To train our LPD network, we use RFBNet with parameters such as: the input image is resized to 512×512 and pass it to the CNN, 200 iterations and mini-batch size of 64, Non-Maximal Suppression(NMS) overlap threshold (IoU) as 0.50, and learning rate of 103 for the first 20 iterations, and 104 after that. It shows 86\% LP detection precision, which concludes that the proposed algorithm can correctly detect an LP from the whole image. The inference time in FPS(frame per second) to detect the LP on NVidia Tesla P4 is shown in Table 4 and on NVidia Jetson TX2 is shown in Table 5.

\begin{table*}[ht]
\centering
\begin{tabularx}{\textwidth}{|c|X|X|X|X|X|X|X|}
\hline
\textbf{Methods} & \textbf{Precision} & \textbf{FPS on Batch Size 1} & \textbf{FPS on Batch Size 2} & \textbf{FPS on Batch Size 4} & \textbf{FPS on Batch Size 8} & \textbf{FPS on Batch Size 16} & \textbf{FPS on Batch Size 32} \\ 
\hline
RFBNet & 86.1\% & 18 & 19 & 20 & 25 & 26 & 56 \\ 
\hline
RFBNet-TensorRT & 86.1\% & 26 & 20 & 22 & 21 & 26 & 60 \\ 
\hline
RFBNet-INT8 & 85.82\% & 50 & 25 & 24 & 20 & 25 & 64 \\ 
\hline
\end{tabularx}
\caption{Results for LPD on NVidia Tesla P4}
\end{table*}

\begin{table*}[ht]
\centering
\begin{tabularx}{\textwidth}{|c|X|X|X|X|X|}
\hline
\textbf{Methods} & \textbf{Precision} & \textbf{FPS on Batch Size 1} & \textbf{FPS on Batch Size 2} & \textbf{FPS on Batch Size 4} & \textbf{FPS on Batch Size 8} \\ 
\hline
RFBNet & 86.1\% & 3.0 & 3.1 & 4.1 & 4.3 \\ 
\hline
RFBNet-TensorRT & 86.1\% & 6.5 & 7.2 & 7.5 & 7.5 \\ 
\hline
RFBNet-FP16 & 86.08\% & 3.2 & 3.9 & 4.4 & 4.4 \\ 
\hline
\end{tabularx}
\caption{Results for LPD on NVidia Jetson TX2}
\end{table*}

\begin{table*}[ht]
\centering
\begin{tabularx}{\textwidth}{|c|X|X|X|X|}
\hline
\textbf{Number of Lines} & \textbf{Total Images} & \textbf{Accuracy} & \textbf{Correct} & \textbf{Incorrect} \\ 
\hline
Single & 1439 & 66.9\% & 981 & 451 \\ 
\hline
Double & 699 & 88.8\% & 1315 & 584 \\ 
\hline
\textbf{Total} & \textbf{2138} & \textbf{94.9\%} & \textbf{1403} & \textbf{627} \\ 
\hline
\end{tabularx}
\caption{CR Results for Recognizable LP of Our Dataset}
\end{table*}

To evaluate the character recognition task, we perform our CR method based on the network shown in Fig. 4 on our dataset for recognizable LP images considering IoU of 0.5 to segment the characters. After that, we filter the results based on our heuristic rules for Singapore LP as discussed in Section 3.3. Our CR method combining our network and the heuristics achieves 67\% of accuracy for recognizing all the characters correctly. It achieves 89\% and 95\% accuracy when considering all the characters are correct except one and two incorrect characters respectively (partial match). Table 6 shows the results of our CR method and it successfully performed segmentation on the majority of the characters. This step takes 68 FPS on NVidia 1080Ti GPU, which shows real-time performance. Since our dataset is composed of Singapore license plates, we could not find a benchmark that would allow us to compare our performance for ALPR in the wild. Comparing across regions, \cite{BrazilLP} has 63.18\% accuracy in the wild. \cite{chinaalpr} achieves a remarkable accuracy of 99.5\% on chinese license plates. Both of these apply to single-line license plates. For Korean license plates \cite{korea} provides a benchmark as well.

\section*{Conclusion}

In this paper, we proposed an end-to-end real-time ALPR system for Singaporean license plates, which are single and double-line, based on Deep Learning approaches. Two CNN-based networks were created: the first (RFBNet) detects license plates and the second (YOLO based) detects and recognizes characters within a cropped license plate image. This system can be used for challenging data with unconstrained capture scenarios, where the LP might be considerably distorted due to oblique views. We evaluated our approach on a dataset consisting of more than 16,000 samples, achieving up to 86\% detection accuracy and 67\% accuracy of the test set for recognizing all the characters correctly. It achieves 89\% and 95\% accuracy when considering all the characters are correct except one and two incorrect characters respectively(partial match). We showed that the proposed system can perform inference in real-time on a variety of hardware architectures including embedded AI computing devices(like Jetson TX2, Xavier).
The proposed method applies to license plates of all regions other than the post-processing module. Post-processing is highly region-specific. Trying to learn the framework of the license plate character distribution from the dataset is quite futile and is better handled in a separate customizable module as we have added.

For future work, we intend to improve our ALPR system by firstly using different CNN models like Spatial Transforming Networks \cite{22} and Parts-Based Networks \cite{27,28} for character recognition. A good recognizer for letters can significantly boost the final accuracy.

\end{document}